%% file: acl.tex
\pdfoutput=1

\documentclass[11pt]{article}

\usepackage[final]{acl}

\usepackage{times}
\usepackage{latexsym}

\usepackage[T1]{fontenc}

\usepackage[utf8]{inputenc}

\usepackage{microtype}

\usepackage{inconsolata}

\usepackage{graphicx}
\usepackage{filecontents}
\usepackage{amsmath,amssymb,amsfonts}
\usepackage{graphicx}
\usepackage{textcomp}
\usepackage{xcolor}
\usepackage{graphicx,verbatimbox,listings,caption,float,lipsum,amsmath,amssymb,algorithm,algpseudocode}
\title{LARA: Linguistic-Adaptive Retrieval-Augmentation for Multi-Turn Intent Classification}

\author{
 \textbf{Junhua Liu\textsuperscript{1,3,$^*$}},
 \textbf{Yong Keat Tan\textsuperscript{2,$^*$}},
 \textbf{Bin Fu\textsuperscript{2,$^\dagger$}},
 \textbf{Kwan Hui Lim\textsuperscript{3}}
\\
\\
 \textsuperscript{1}Forth AI
 \\
 \textsuperscript{2}Shopee
 \\
 \textsuperscript{3}Singapore University of Technology and Design
}

\begin{document}
\maketitle
\def\thefootnote{*}\footnotetext{Equal Contributions.}\def\thefootnote{\arabic{footnote}}
\def\thefootnote{$\dagger$}\footnotetext{Correspondence: \texttt{bin.fu@shopee.com}}\def\thefootnote{\arabic{footnote}}

\input{Sections/Abstract}

\input{Sections/Introduction}
\input{Sections/RelatedWork}
\input{Sections/Problem}

\input{Sections/Solution}
\input{Sections/Experiments}

\input{Sections/Results}
\input{Sections/Ablation}
\input{Sections/Limitation}
\input{Sections/Conclusion}

\bibliography{ieee}

\appendix


\input{Sections/Appendix}

\end{document}

%% file: Sections/Abstract.tex
\begin{abstract}

Multi-turn intent classification is notably challenging due to the complexity and evolving nature of conversational contexts. This paper introduces LARA, a Linguistic-Adaptive Retrieval-Augmentation framework to enhance accuracy in multi-turn classification tasks across six languages, accommodating a large number of intents in chatbot interactions. LARA combines a fine-tuned smaller model with a retrieval-augmented mechanism, integrated within the architecture of LLMs. The integration allows LARA to dynamically utilize past dialogues and relevant intents, thereby improving the understanding of the context. Furthermore, our adaptive retrieval techniques bolster the cross-lingual capabilities of LLMs without extensive retraining and fine-tuning. Comprehensive experiments demonstrate that LARA achieves state-of-the-art performance on multi-turn intent classification tasks, enhancing the average accuracy by 3.67\% from state-of-the-art single-turn intent classifiers.


\end{abstract}

%% file: Sections/Introduction.tex
\section{Introduction}

A chatbot is an essential tool that automatically interacts or converses with customers. It plays a crucial role for international e-commerce platforms due to the rising consumer demand for instant and efficient customer service. Chatbots represent a critical component of dialogue systems \cite{weld2021survey} that can answer multiple queries simultaneously by classifying intent from the user's utterance to reduce waiting times and operational costs. Naturally, the interaction with users could turn into a multi-turn conversation if they require more detailed information about the query. Developing an intent classification model for a dialogue system is not trivial, even if it is a typical text classification task. As we must consider contextual factors such as historical utterances and intents, failing to understand session context while recognizing user intention often leads to visible errors. It would invoke a completely wrong application or provide an unrelated answer \cite{6853573}. As a result, it faces several challenges in dialogue understanding.  

\begin{figure*}[t]
    \centering
    \includegraphics[width=.95\textwidth]{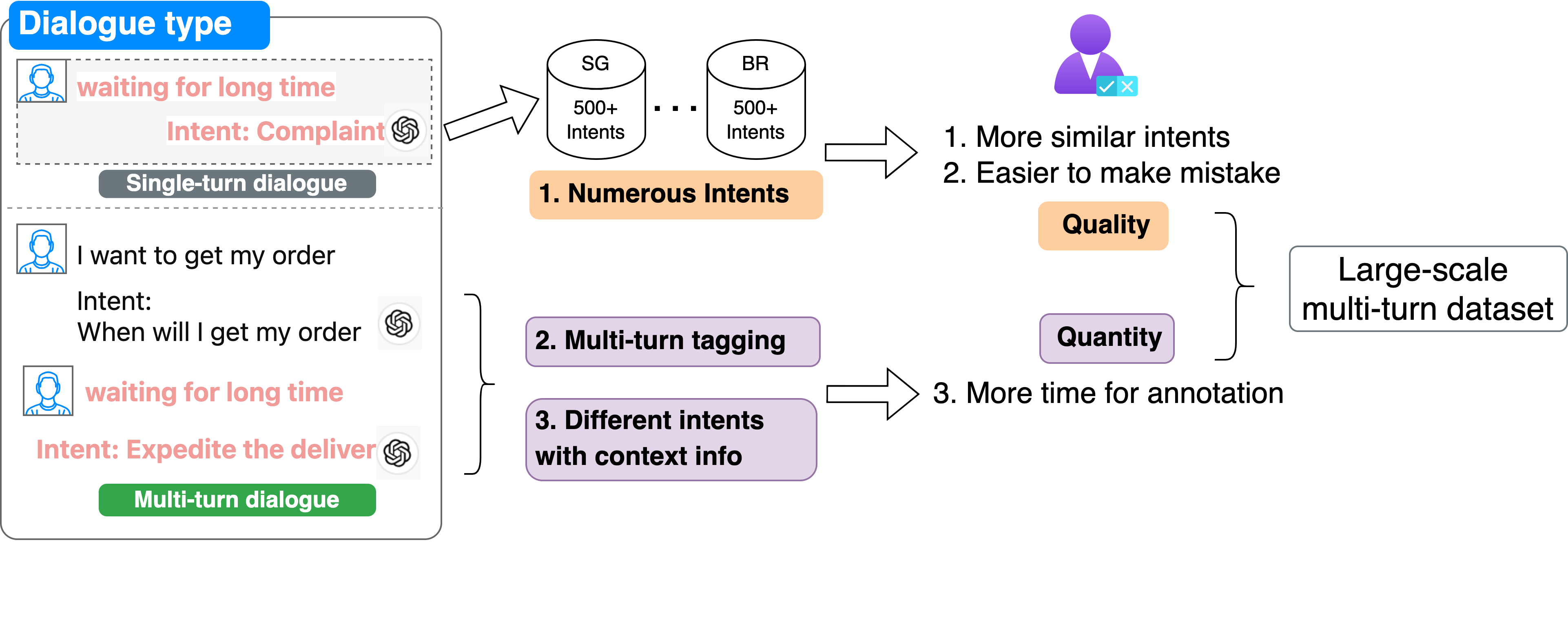}
    \vspace{-1cm}
    \caption{Annotation pipeline of multi-turn intent classification dataset }
    \label{fig:challenges}
    \vspace{-0.3cm}
\end{figure*}

The biggest challenge is that multi-turn datasets are hard to collect. 
While there are some studies on dialogue understanding in multi-turn intent classification~\cite{9082602,wu2021contextaware,Qu_2019}, they are made under the assumption of the availability of multi-turn training data, which is usually not the case in the real world. 

Figure~\ref{fig:challenges} shows the annotation pipeline of multi-turn intent classification. Unlike emotion recognition in conversation (ERC) with only less than 10 classes or topic classification within dialogue state tracking (DST) with tens of topics, there are hundreds of intents within the knowledge base of a chatbot to cover users' specific intents in each market, which increases the complexity of classification tasks and multi-turn data annotation. Annotators can easily make mistakes and spend more time making decisions due to the numerous intents. Combined, these make it a high-cost and time-consuming annotation task, and it is unrealistic to annotate large-scale multi-turn datasets manually. However, the performance will most likely suffer without enough training sample size. This calls for a more efficient method in solving the challenge \cite{Mo2023LearningTR}. 



To tackle the above challenge, we propose \textbf{L}inguistic-\textbf{A}daptive \textbf{R}etrieval-\textbf{A}ugmentation, or LARA, which offers a pipeline of techniques to adopt only single-turn training data to optimize multi-turn dialogue classification. LARA first leverages an XLM-based model trained on single-turn classification datasets for each market, thus simplifying data construction and maintenance. Subsequently, LARA advances the field by selecting plausible candidate intents from user utterances and employing a retriever to gather relevant questions for prompt construction. This process facilitates in-context learning (ICL) with multi-lingual LLMs (MLLMs), significantly enhancing model efficacy without requiring market-specific multi-turn models.


In summary, the contributions of this paper are as follows:


\begin{enumerate}
  \item We introduce LARA as a multi-turn classification model only leveraging single-turn datasets to effectively address multi-turn data collection issues. 
  \item We conduct experiments on our e-commerce multi-turn dataset across six languages. LARA achieves state-of-the-art results and reduces inference time during ICL with MLLMs.
\end{enumerate}

%% file: Sections/RelatedWork.tex
\section{Related Work}

\textbf{Modeling Multi-turn Dialogue Context}: Modelling the multi-turn dialogues is the foundation for dialogue understanding tasks. Previous works adopt bidirectional contextual LSTM \cite{ghosal2021exploring,liu2022title2vec} to create context-aware utterance representation on MultiWOZ intent classification \cite{budzianowski2018multiwoz}. Recent works use PLM as a sentence encoder \cite{shen2021directed} on emotion recognition in conversation (ERC). Specifically, \cite{lee2022compm} used PLM to encode the context and speaker's memory and \cite{qin2023bert} enhance PLM by integrating multi-turn info from the utterance, context and dialogue structure through fine-tuning. However, all of their tasks adopt the multi-turn dialogue training set, which is hard to collect for an e-commerce chatbot. Our method attempts to combine an XLM-based model trained on the single-turn dataset into an in-context retrieval augmented pipeline with LLM, solving the multi-turn intent classification task in a zero-shot setting. 


\noindent\textbf{In-context Retrieval}: In-context learning (ICL) with LLM like GPT-3 \cite{gpt3} demonstrates the significant improvement on few-shot/zero-shot NLP tasks. ICL has been successful in utterance-level tasks like intent classification \cite{Yu2021FewshotIC}. As for the \textbf{Retrieval} part, most research on in-context learning (ICL) usually deals with single sentences or document retrieval, but we are interested in finding and understanding dialogues. Generally, there are two types of systems to find the relevant dialogues: the first is LM-score based retrieval. They \cite{Rubin2021LearningTR, Shin2021ConstrainedLM} check the probability of a language model, like GPT-3, to decode the right answer based on an example. The second type defines similarity metrics between task results and uses them as the training objective for the retriever. Both K-highest and lowest examples are used as positive and negative samples to help the system learn. The most pertinent research on dialogue retrieval concentrates on areas such as knowledge identification \cite{Wu2021DIALKIKI} and response selection \cite{Yuan2019MultihopSN}. Our objectives and settings differ from them.

%% file: Sections/Problem.tex
\section{Problem Formulation}




\subsection{Hierarchical Text Classification}
Hierarchical text classification (HTC) is a type of text classification where the classes or categories are organized in a hierarchy or tree structure. Instead of having a flat list of categories, the categories are arranged in a nested manner, so we need to consider the relationships of the nodes from different levels in the class taxonomy. 

The intents in our scenario are organised in a hierarchical tree structure. Specifically, each category can belong to at most one parent category and can have arbitrary number of children categories. Our class taxonomy $\mathcal{T}$ is a tree with fixed depth 3 and one meta root node, that is, the distance from the root node to all leaf nodes is 3, and the unique path from the root node to each leaf node will form one intent. 

We formulate the single-turn HTC task in our scenario as such, given a text $q_i$, the purpose is to predict a subset $I$ of the complete intent set $\mathcal{I}$, where the size of the subset $|I|$ containing the category from each layer is 3 (excluding root node). Generally, the number of intent sets exceeds 200.  


\subsection{Multi-turn Intent Classification} Multi-turn scenario shares the same $\mathcal{T}$ and $\mathcal{I}$ as single-turn scenario. It involves a series of user queries $\mathcal{Q} = \{q_i\}_{i=1}^n$ in dialogues, the objective is to identify the intent of the final query $q_n$. Multi-turn recognition must account for the entire conversational context $\mathcal{C} = \{q_i\}_{i=1}^{n-1}$, which includes the historical queries. This context-dependency introduces additional complexity, requiring models to interpret nuanced conversational dynamics and adjust to evolving user intentions over the course of an interaction.

\subsection{Objective} This work aims to use easily accessible single-turn data to improve the accuracy of multi-turn intent recognition without requiring any multi-turn training datasets.

%% file: Sections/Solution.tex
\begin{figure*}[t]
    \centering
    \includegraphics[width=.9\textwidth]{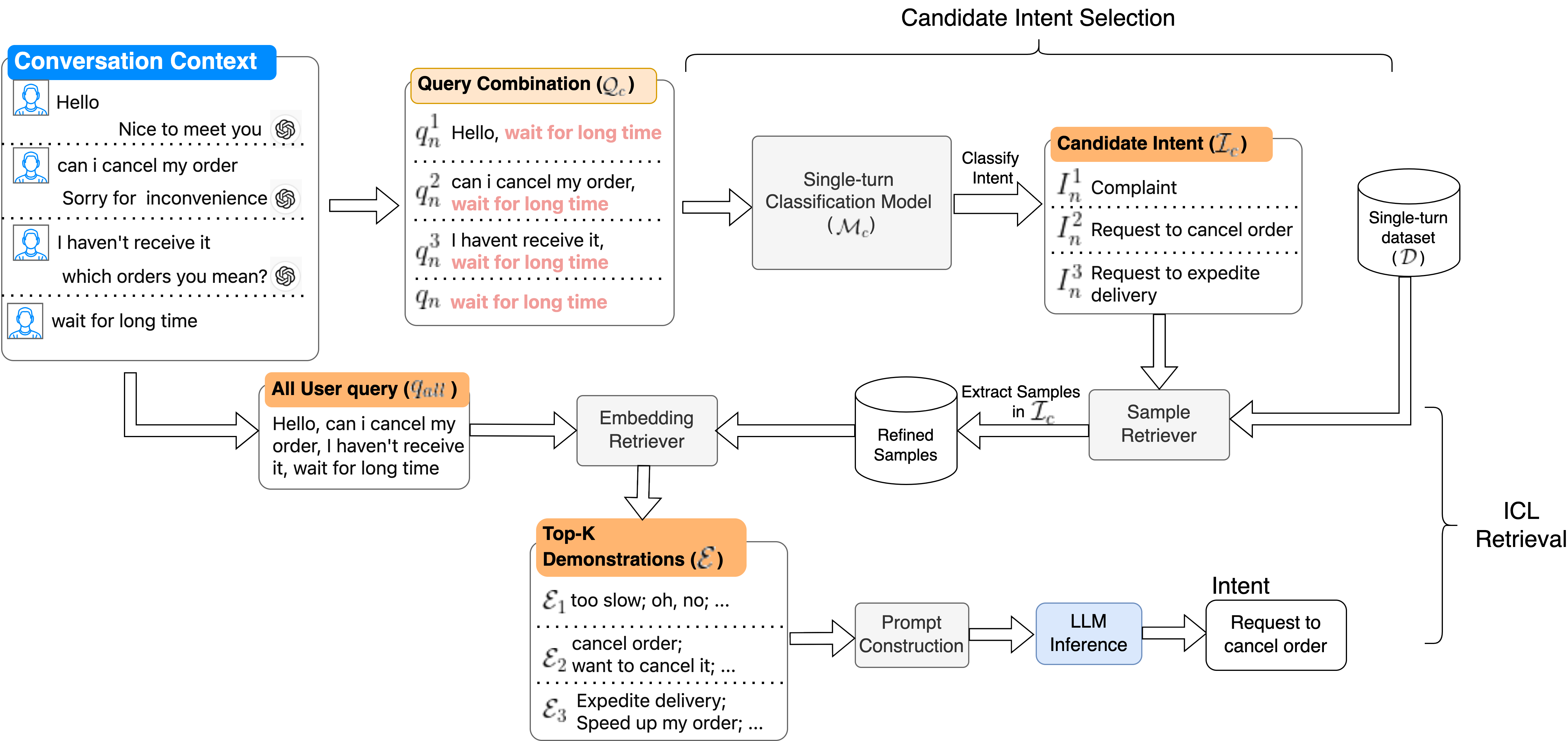}
    \caption{The pipeline of Linguistic-Adaptive Retrieval-Augmentation}
    \label{fig:ICLway}
    \vspace{-0.2cm}
\end{figure*}

\section{LARA: Linguistic-Adaptive Retrieval-Augmentation}

The LARA framework shown in Figure \ref{fig:ICLway} addresses the multi-turn intent recognition challenge through zero-shot in-context learning with single-turn demonstrations guided by a crafted instruction prompt. First, a single-turn classification model $\mathcal{M}_c$ is trained on single-turn dataset and used to narrow down the intents to be included in the ICL prompt, which are henceforth referred to as candidate intents. This step is necessary due to the limited LLM context window, and it also helps to filter out extra noise from direct demonstration retrieval. Then, for every candidate intent, in-context demonstrations are selected by retrieving single-turn examples that are semantically similar to the multi-turn test sample. Finally, an instruction prompt for multi-turn classification is formulated by combining the demonstrations and test user queries.

\subsection{Single-turn Classification Model ($\mathcal{M}_c$)}
Before diving into LARA, we must train a single-turn hierarchical text classification model on our single-turn dataset $\mathcal{D}$. The model is an ensemble of a simple label-attention model \cite{zhang2020multi} and a state-of-the-art HTC approach named HiTIN \cite{Zhu2023HiTINHT}. The label-attention model only considers the semantic info of user utterances and label-query attention info, which ignores the hierarchical tree-like structure in our intent system. So, we build model $\mathcal{M}_c$, integrating taxonomic structure via the tree isomorphism network within HiTIN into our label-attention model. More details are provided in Appendix \ref{appendix:a}.

We conduct some experiments on our multilingual dataset. The result in Table \ref{table:htc_improvement} shows that HiTIN alone performs better than the label-attention model, and ensembling the two methods further improves performance on average scores. We will use this model as our baseline and generate candidate intents within LARA.


\begin{table*}[t]
\centering
\scalebox{0.8}{
\begin{tabular}{||l c c c c c c c c c||} 
 \hline
  & BR & ID & MY & PH & SG & TH & TW & VN & avg \\ [0.5ex] 
 \hline\hline
 Traffic weight & 150k & 212k & 27k & 46k & 5k & 36k & 36k & 43k &  \\
 \hline\hline
 Label-attention model & 81.55\% & 87.91\% & 83.73\% & 73.29\% & 83.69\% & 83.11\% & 71.55\% & 74.44\% & 82.30\% \\
 HiTIN & 81.85\% & 89.41\% & 84.57\% & 74.75\% & 84.17\% & 83.67\% & \textbf{75.38}\% & \textbf{75.99}\% & 83.53\% \\
 Ensembled  & \textbf{82.66}\% & \textbf{89.62}\% & \textbf{85.33}\% & \textbf{75.98}\% & \textbf{84.57}\% & \textbf{83.89}\% & 75.19\% & 75.52\% & \textbf{83.94}\% \\[1ex]
 \hline
\end{tabular}}
\caption{Noticeable improvement from adding a state-of-the-art HTC approach.} 
\vspace{-0.2cm}
\label{table:htc_improvement}
\end{table*}

\subsection{Candidate Intents Selection}
\textbf{Query Combination}: After receiving the context from user side, the last query $q_n$ is first combined with each historical query in conversational context $\mathcal{C}$ to form a query combination set $\mathcal{Q}_c$ = $\{q_n, q^1_n, ..., q^{n-1}_n\}$, where $q^i_n$ means the text concatenation of $q_i$ with $q_n$ using a comma. 

\noindent\textbf{Candidate Intent Recognition}: $\mathcal{M}_c$ predicts few candidate intents $\mathcal{I}_{c}$ from all intents $\mathcal{I}$ on these query combinations $\mathcal{Q}_c$. The $\mathcal{M}_c$ is then used to perform inference on each of the concatenated queries to get a set of candidate intents $\mathcal{I}_{c} = \{I_{q_n}, I_{q^1_n}, ..., I_{q^{n-1}_n}\}$. Finally, we will select the top 3 intents with the highest scores. Note that $\mathcal{I}_{c}$ is a set, so duplicated intents will be removed, and the maximum size of $\mathcal{I}_{c}$ is equal to the number of queries $n$ in a session.



\subsection{ICL Retrieval}

However, not all training examples under the candidate's intent $\mathcal{I}_{c}$ will be used in the prompt as demonstrations. Here, the demonstrations refer to a sequence of annotated examples that provide LLM with decision-making evidence and specify an output format for natural language conversion into labels during ICL. Our strategy is to sample training examples that are similar to the test sequence for each candidate intent, a method introduced by \cite{liu2021makes}. In this process, we first concatenate all the queries in the session with a comma to get the test sequence $q_{all}$. Then, $q_{all}$ is mapped to a vector $H_{q}$ using the [CLS] token embedding from a pre-trained sentence encoder, $\Phi_{XLMR}$. After that, we retrieve all the single-turn training samples for each candidate intent from $\mathcal{D}$ using a sample retriever and also encode them using $\Phi_{XLMR}$. With $H_{q}$ as query, we use an embedding retriever to search through the sampled examples to retrieve top $K$ - 1 nearest data examples of each candidate intent based on their cosine similarity to $H_{q}$. Together with the representative query of each candidate intent, the retrieved single-turn samples of all candidate intents form the demonstrations $\mathcal{E}$ for in-context learning. We show the detailed algorithm in \ref{alg:retrieval} below.

\subsection{Prompt Construction and LLM Inference}
A task instruction $T$ is hand-crafted to guide the model to perform multi-turn intent recognition task by referring to the single-turn demonstrations. The task instruction $T$, combined with demonstrations $\mathcal{E}$, conversational context $\mathcal{C}$, and the query $q_n$, forms the input prompt $\mathcal{P}$ for the LLM. The concatenation of each prompt component into one long text is shown in the appendix \ref{appendix:c}. To accommodate real-time application latency requirements, two additional methods were explored to constrain the model to generate single-token symbols representing intents, detailed as $\mathcal{P}_{symbolic}$ and $\mathcal{P}_{prepend}$, with examples also provided in the appendix. Model outputs are greedily decoded, ensuring efficient and accurate intent recognition.



\begin{algorithm}[t]
\caption{ICL Demonstrations Retrieval}\label{alg:retrieval}
\begin{algorithmic}
\let\oldReturn\Return
\renewcommand{\Return}{\State\oldReturn}
\Require $\mathcal{I}_c$, $Q$, a positive integer $K$
\State $ \mathcal{I}' = remove\_duplicate(\mathcal{I}_c) $
\State $ q_{all} = text\_concatenate(\mathcal{Q}) $
\State $ H_q = \Phi_{XLMR}(q_{all})^{[CLS]} $
\State
\For { each item $I_i$ in $\mathcal{I}'$ }
  \State /* Sample retriever: Get the single-turn training samples under intent $I_i$ */
  \State $ X_i = sample\_from\_\mathcal{D}\_for\_intent(I_i) $
  \State
  \State /* Embedding retriever: Get embedding for each training sample */
  \State $ H_{X_i} = \{\Phi_{XLMR}(x_j)^{[CLS]}\}_{j=1}^{|X_i|}, x \in X_i $
  \State
  \State /* Calculate text similarity of each training sample with test queries */
  \State $ S_i = \{cosine\_similarity(H_q, h_j)\}_{j=1}^{|H_{X_i}|}$, where $h \in H_{X_i} $
  \State
  \State /* Get top $K$ nearest demonstrations */
  \State $ \mathcal{E}_i \gets $ Top $ (K-1) $ $ x \in X_i$ based on $S_i$
  \State Append $r$ of $I_i$ to $\mathcal{E}_i$ /* add representative query to demonstrations of $I_i$ */
\EndFor
\State
\State $ \mathcal{E} = \{\mathcal{E}_i\}_{i=1}^{|\mathcal{I}'|} $ /* collect demonstrations of all candidate intents */
\State $ S = \{S_i\}_{i=1}^{|\mathcal{I}'|} $
\State Sort $ \mathcal{E} $ by their scores $S$ in ascending order
\State
\Return $ \mathcal{E} $
\end{algorithmic}
\end{algorithm}

%% file: Sections/Experiments.tex
\section{Experiments}
\subsection{Experimental Setup}
\textbf{Dataset}. The dataset is obtained from the conversation history of a large e-commerce company. It consists of user queries in the local languages of eight markets: Brazil, Indonesia, Malaysia, Philippines, Singapore, Thailand, Taiwan, and Vietnam.  All the labelled data are collected through manual annotation by local customer service teams of each market. The samples with consistency labels from 3 taggers are selected to ensure the annotation quality. More details are in Appendix \ref{appendix:b}.

\noindent\textbf{Metrics}. We evaluate the accuracy of the methods based only on the label of the last query $q_n$ in each conversation session $\mathcal{Q}$. Other metrics which consider class imbalance are not used as the sampled sessions are expected to reflect the online traffic of each intent, thus more accurately simulating true online performance.

\subsection{Baselines}
Current multi-turn models are trained with multi-turn datasets, but our methods did not require such data. For a fair comparison, we adopt a state-of-the-art single-turn model mentioned in 4.1 with two types of concatenation approaches as baselines:
\subsubsection{Single-turn approach}
Inference on only the last utterance of a session using the single-turn model $\mathcal{M}_c$, all previous contexts are ignored.
\subsubsection{Naive concatenation}
All queries in a single session $\mathcal{Q}$ are concatenated using comma, and the concatenation result is fed into the single-turn model $\mathcal{M}_c$, which is enhanced by HTC approach mentioned above, for inference. HTC methods which optimally utilise the overall label hierarchy information often outperform the methods which simply disregard the structure\cite{Rojas2020EfficientSF}. 
\subsubsection{Selective concatenation}
In this approach, only one query from $\mathcal{C}_q$ is selected to be concatenated with $q_n$. The intuition is that not all history queries are helpful in understanding the last query, and the excessive use of them might introduce unwanted noise. A concatenation decision model is trained to select the most appropriate historical query. Depending on the model confidence, there might be cases where no expansion is needed at all. The concatenation result is then also fed into the single-turn model $\mathcal{M}_c$ for inference.

%% file: Sections/Results.tex
\section{Results and Discussions}
Table \ref{table:results_icl} compares the performance of baselines and LARA on our multi-turn dataset. The single-turn approach has the worst performance due to the lack of context from history queries. The single-turn model with \emph{Naive concatenation} is lower than \emph{Selective concatenation} by 0.89\% on average, showing that naively including all history queries will introduce noises, which in turn jeopardizes the performance. However, pseudo-labelling the dataset used to train the concatenation decision model will need to be carefully carried out, and despite the extra steps, it will not necessarily be more effective than the naive method. 

LARA, on the other hand, with prompt $\mathcal{P}_{formatted}$, achieves the best results on most markets without any multi-turn training data. On average, it improved accuracy by 3.67\% compared with \emph{Selective concatenation}. Especially on the non-English markets, it also improved by 2.96\%, 4.18\% and 4.00\% on TH, VN and BR separately. This highlights the linguistic-adaptivity of the method on broad languages. The only market that does not outperform the baselines is ID, which most probably can be attributed to the language ability of open-sourced LLMs in handling the local slang and abbreviations in casual conversation. After all, the backbone model used in baselines is pre-trained directly on the in-domain chat log data, while the LLM models are used out of the box.

\begin{table*}[th!]
\centering
\scalebox{.8}{
\begin{tabular}{||l l c c c c c c c c c||} 
 \hline
 Model & Prompt & BR & ID & MY & PH & SG & TH & TW & VN & avg \\ [0.5ex] 
 \hline\hline
 Single-turn & - & 30.98\% & 52.14\% & 56.81\% & 40.21\% & 51.13\% & 52.99\% & 58.07\% & 65.90\% & 53.76\% \\
 Naive Concat. & - & 50.81\% & 60.61\% & 57.02\% & 47.62\% & 60.52\% & 56.97\% & 65.44\% & 76.95\% & 60.08\% \\
 Selective Concat. & - & 52.69\% & \textbf{63.23}\% & 60.20\% & 51.32\% & 56.99\% & 57.77\% & 64.02\% & 74.10\% & 60.97\% \\ 
 Vicuna-13B & $\mathcal{P}$ & 52.69\% & 61.48\% & \textbf{65.42}\% & \underline{54.50}\% & 65.26\% & 60.96\% & \textbf{67.14}\% & \underline{77.90}\% & \underline{64.18}\% \\
 Vicuna-13B & $\mathcal{P}_{symbolic}$ & 51.88\% & 60.00\% & 64.57\% & 53.97\% & 65.26\% & 58.96\% & 65.44\% & 74.67\% & 62.92\% \\
 Vicuna-13B & $\mathcal{P}_{prepend}$ & \underline{54.03}\% & 61.75\% & 64.50\% & 53.44\% & \textbf{65.94}\% & \underline{61.55}\% & \underline{66.86}\% & 75.81\% & 63.97\% \\
 Vicuna-13B & $\mathcal{P}_{formatted}$ & \textbf{55.65}\% & \underline{62.88}\% & \underline{64.71}\% & \textbf{55.03}\% & \underline{65.40}\% & \textbf{61.95}\% & \underline{66.86}\% & \textbf{78.10}\% & \textbf{64.64}\% \\[1ex]
 \hline
\end{tabular}
}
\caption{Performance of LARA compared to baselines, the average here is weighted on the number of test samples in each market. The best performance for each dataset is in boldface, while the second best is underlined.}
\label{table:results_icl}
\vspace{-0.3cm}
\end{table*}

Replacing the label names with non-related symbols in $\mathcal{P}_{symbolic}$ significantly hurts the performance of in-context learning. On the other hand, minimal changes to label names in $\mathcal{P}_{prepend}$ does not heavily impact the performance. In turn, the inference time is improved by 77\%, from 0.75it/s to 1.32it/s on a single V100 card using Hugging Face python library. Interestingly, the model also stopped generating labels which cannot be matched with the options provided in demonstrations, while previously the rate is on average 1.6\% using $\mathcal{P}$. Finally, we also tried a new prompt $\mathcal{P}_{formatted}$ based on $\mathcal{P}_{prepend}$. Only a very slight change to the context format is done, but it can outperform the other prompt variants in all datasets, suggesting that giving $\mathcal{C}_{\mathcal{Q}}$ a closer format to $\mathcal{E}$ and the targeted $q_n$ will be more beneficial in the context utilization. Besides, this also hints that the prompt could also be worked on more in the future as it is not extensively tuned in this work. 

%% file: Sections/Ablation.tex
\begin{table*}[t!]
\centering
\scalebox{.8}{
\begin{tabular}{||l c c c c c c c c c||} 
 \hline
 Encoder & BR & ID & MY & PH & SG & TH & TW & VN & avg \\ [0.5ex] 
 \hline\hline
 Our own encoder & \underline{55.65\%} & \textbf{62.88\%} & \underline{64.71\%} & 55.03\% & \textbf{65.40\%} & \underline{61.95\%} & 66.86\% & \textbf{78.10\%} & \textbf{64.64\%} \\
 
 all-MiniLM-L12-v2 & 54.57\% & 61.05\% & 64.36\% & 53.97\% & \underline{65.13\%} & 59.56\% & 68.27\% & 75.81\% & 63.63\% \\

 all-mpnet-base-v2 & \textbf{55.91\%} & 60.79\% & 63.30\% & 54.50\% & 64.31\% & 58.57\% & 67.42\% & \underline{76.95\%} & 63.24\% \\

 paraphrase-multilingual & & & & & & & & & \\
 -MiniLM-L12-v2 & 54.03\% & \underline{62.01\%} & 64.42\% & \textbf{57.14\%} & 64.31\% & \textbf{62.35\%} & \textbf{69.12\%} & \underline{76.95\%} & 64.25\%\\

 paraphrase-multilingual & & & & & & & & & \\
 -mpnet-base-v2 & 55.38\% & 61.57\% & \textbf{65.42\%} & \underline{55.56\%} & 64.99\% & 59.96\% & \underline{68.56\%} & \underline{76.95\%} & \underline{64.29\%} \\[1ex]
 \hline
\end{tabular}
}
\caption{Performance of LARA using different encoders for in-context demonstration mining. The best performance for each dataset is in bold, while the second best is underlined.}
\label{table:results_encoder}
\vspace{-0.3cm}
\end{table*}

\section{Ablation Studies}
To validate our motivation and model design, we ablate single-turn model $\mathcal{M}_c$ in candidate intent selection and retrievers in ICL retrieval. The comparison is made on the original $\mathcal{P}$ prompt variant.

\subsection{The necessity of model $\mathcal{M}_c$}
$\mathcal{M}_c$ is used to recognise the intent candidates before ICL retrieval. If so, all demonstrations are directly retrieved based on their cosine similarity to $q_{all}$ and the quality of in-context learning is adversely impacted. The accuracy on all markets dramatically dropped except PH, which only dropped by 0.53\% due to the least number of intents. The average score across all markets dropped from 64.18\% to 53.99\%. For instance, "refund timeline" and "refund timeline for cancelled order" could be confusing to retrieval-based models, while classification models trained on each market dataset can discern them better.

\begin{figure}[t]
    \centering
    \includegraphics[width=.48\textwidth]{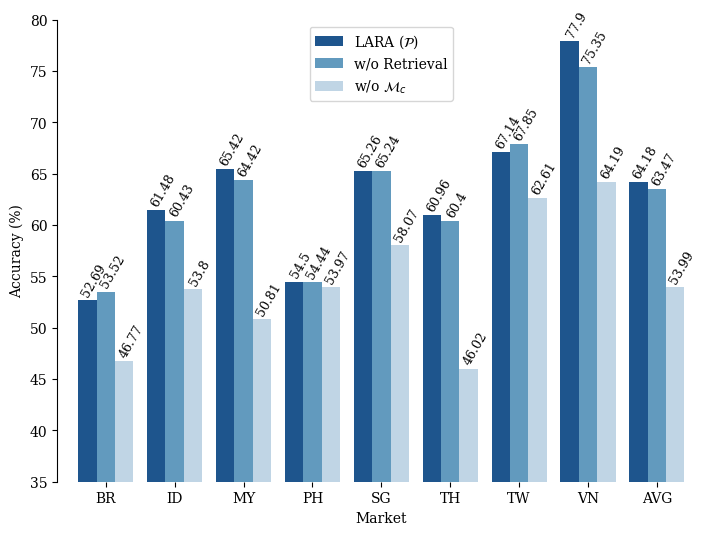}
    \vspace{-1cm}
    \caption{Ablation on different components of LARA.}
    \label{fig:ablation}
    \vspace{-0.3cm}
\end{figure}

\subsection{The role of retrievers in ICL retrieval}
Demonstration selection via retrievers may have an impact on the performance. Thus, we remove all retrievers and randomly sample the demonstrations for each intent. The results are reported with 10 runs on the random sampling. As shown in Figure $\ref{fig:ablation}$, the overall performance decreased by 0.71\% without retrievers, highlighting the importance of selecting demonstrations which are more similar to test queries. ID and VN, with demonstration pool two to three times bigger than others (refer to Appendix \ref{appendix:b} for pool size), are affected the most because the chance of selecting samples which are not so similar is higher.

\subsection{Quality of Embedding Model}
We also experimented with open-source, multilingual semantic similarity models provided by SentenceTransformers \cite{reimers-2019-sentence-bert} for in-context demonstration mining. These models are easily accessible to the public and cover a wide range of languages. Due to the training method, the similarity metric used is still cosine similarity.

The multilingual models selected in this experiment have lower or roughly the same number of parameters as our own XLM-RoBERTa-base model used in the paper. They are
\begin{itemize}
   \item \textbf{Models with all-rounded abilities} which have the best average performance reported by the authors: \textit{all-MiniLM-L12-v2} and \textit{all-mpnet-base-v2}
   \item \textbf{Models for paraphrase mining} as it is similar to our task of comparing multi-turn utterances to single-turn utterances: \textit{paraphrase-multilingual-MiniLM-L12-v2} and \textit{paraphrase-multilingual-mpnet-base-v2}
\end{itemize}

All comparisons are done using the same prompt, $\mathcal{P}_{formatted}$. From the table \ref{table:results_encoder}, the best open-source model of selections (from row 2 - 5) is overall only 0.35\% behind our own model, which means that the quality of the embedding model does not matter much and can easily be replaced by open-source models. That said, we would recommend paraphrase mining models over the general purpose ones as they are more suited for the scenario. Furthermore, if resource is a concern, smaller models like \textit{MiniLM-L12} can be selected with 3x speed up compared to \textit{mpnet-base}, all while maintaining the same overall quality.

%% file: Sections/Limitation.tex
\section{Limitation}
LARA does not currently address the detection of out-of-domain utterances, a critical aspect for online dialogue systems. Future research is necessary to explore methods for incorporating this capability and to assess their feasibility. Furthermore, the resilience of the method to user intent shifts has not been examined. Additionally, the multi-component architecture, which integrates text classification, retrieval, and ICL, adds to the implementation complexity. In Appendix \ref{apendix:d}, we propose a lightweight solution that is more suitable for applications with limited deployment resources.

%% file: Sections/Conclusion.tex
\section{Conclusion}
This paper introduced LARA, a framework that leverages Linguistic-Adaptive Retrieval-Augmentation to address multi-turn intent classification challenges through zero-shot settings across multiple languages. Unlike other supervised Fine-Tuning (SFT) models, which require a hard-to-collect multi-turn dialogue set, our method requires only a single-turn training set to train a conventional model, combining it with an innovative in-context retrieval augmentation for multi-turn intent classification. LARA substantially improved user satisfaction by 1\% over multi-turn sessions, reduced the transfer-to-agent rate by 0.5\% and saved the cost of hundreds of agents, which is a key business metric in our industrial application.


The empirical results underscore LARA's capability to enhance intent classification accuracy by 3.67\% over existing methods while reducing inference time, thus facilitating real-time application adaptability. Its strategic approach to managing extensive intent varieties without exhaustive dataset requirements presents a scalable solution for complex, multi-lingual conversational systems.

For future work, we intend to extend and apply the LARA framework to recommendation tasks in other domains, such as understanding how user intent may shift during a POI or itinerary recommendation for tourism purposes~\cite{halder2024survey}. This will enable us to better capture evolving user preferences due to temporal shifts, changing contexts, and individual or group behavioural patterns, which is also applicable to general sequence recommendations.

\vspace{3mm}
{
{\noindent\bf Acknowledgments.} 
This research is supported in part by the Ministry of Education, Singapore, under its Academic Research Fund Tier 2 (Award No. MOE-T2EP20123-0015). Any opinions, findings and conclusions, or recommendations expressed in this material are those of the authors and do not reflect the views of the Ministry of Education, Singapore.
}

%% file: Sections/Appendix.tex
\lstnewenvironment{prompt}[1][]{%
  \lstset{
    basicstyle=\ttfamily,
    frame=tb,
    escapeinside=||,
    breaklines=true,
    #1
  }%
}{}

\section{The details of single-turn model ($\mathcal{M}_c$)}
\label{appendix:a}

A text classification model is trained on the annotated single-turn dataset $\mathcal{D}$. The model is an ensemble of a simple label-attention model and a state-of-the-art global HTC approach. The label-attention model exploits local information per layer of the taxonomy by having separate classifier head for each intent taxonomy layer, whereas the global approach addresses the task with a single model for all the classes and levels. In our implementation, both approaches are trained as a single network and back propagation is performed on the ensembled output.

The approaches share the same encoder and sentence representation. Given a query $q$, we adopt the [CLS] token embedding from XLM-RoBERTa-base model with weight $\Phi_{XLMR}$ as the text representation $H$. Formally, 
$$ H = \Phi_{XLMR}(q)^{[CLS]} \in \mathbb{R}^{d} $$ where $d$ is the hidden dimension. 
$\Phi_{XLMR}$ had been further pretrained with our in-domain corpus to give meaningful representation to [CLS] token. 

In the label-attention model, we have one classifier head for each intent layer. Each of the classifier heads has one hidden linear layer to obtain the layer intermediate output $L_l$, which encodes the layer information. This layer information will be utilised in the input of of the next layer classifier head. 

\[
    L_l = 
    \begin{cases}
        HW^1_l + b^1_l, & \text{if } l = 1, \\
        (H \oplus L_{l-1})W^1_l + b^1_l, & \text{if } l > 1,
    \end{cases}
\]
where \( W^1_l \in \mathbb{R}^{d \times d} \text{ for } l = 1 \text{ and } W^1_l \in \mathbb{R}^{2d \times d} \text{ for } l > 1 \). $b^1_l \in \mathbb{R}^d$, $l$ is the layer number, $\oplus$ denotes tensor concatenation. Finally, we obtain the local logits $H_{local}^l$ for each layer classes by using another linear layer
$$ H_{local}^l = L_l \cdot W^2_l + b^2_l, W^2_l \in \mathbb{R}^{d \times |\mathcal{I}_l|}, b^2_l \in \mathbb{R}^{|\mathcal{I}_l|} $$
where $|\mathcal{I}_l|$ is the number of classes in the layer. 

However, the label-attention model used is not aware of the overall hierarchical structure. Therefore, we ensemble it with another method. We refer to HiTIN \cite{Zhu2023HiTINHT} for the implementation of state-of-the-art HTC global approach. In this method, a tree network is constructed based on the simplified original taxonomy structure, and the messages are propagated bottom-up in an isomorphism manner, which complements the label-attention model used. The embedding for leaf nodes are obtained by broadcasting the text representation $H$. After the tree isomorphism network propagation, all embedding from all layers are aggregated to form single embedding, and a classification layer is used to obtain the logits $H_{global}$ of all tree nodes. The logits are then split by the number of classes in each layer to obtain $H_{global}^l$.

The final class probabilities for each layer $P_l$ is then obtained by
$$ P_l = softmax(H_{local}^l + H_{global}^l) $$


\section{Implementation Details}
\label{appendix:b}




The traditional single-turn model, the retriever, and the concatenation decision model used are using backbone initialized with $\Phi_{XLMR}$, a multi-lingual domain specific XLM-RoBERTa-base model continued to be pre-trained with contrastive learning. We use AdamW to finetune the backbone and all other modules with a learning rate of 5e-6 and 1e-3, respectively. In LARA, the LLM used is vicuna-13b-v1.5 on Hugging Face with 13B parameters. All test are run on a single Nvidia V100 GPU card with a 32GB of GPU memory. The number of demonstrations $K$ retrieved for each intent is set at 10 in this experiment. We also experimented with numbers below 10, the lower the number, the lower the performance. 10 is the highest number we can fit due to GPU memory constraint. Due to this constraint as well, the total number of tokens the in-context learning demonstrations can make up to are limited to 2300 tokens. If exceeded, the number of demonstrations in each candidate intent are pruned equally starting with the ones with the lowest cosine similarity scores to $q_{all}$. During inference time, if the generated intent does not match any of the provided options, the intent of $\mathcal{M}_c$ on $q_n$ will be considered as the final result. 

\subsection{Dataset Details}
Table \ref{table:data_train_test} shows the number of samples in each dataset.  We have the single-turn training data available in abundance over the course of business operations after years. These single-turn samples will serve as the demonstration pool for in-context learning. To evaluate the effectiveness of our methods, we also have the CS teams to manually annotate some real multi-turn online sessions to serve as the test set. Each session queries $\mathcal{Q}$ will only have the last query $q_n$ labelled. 
\begin{table}[t]
\centering
\scalebox{0.9}{
\begin{tabular}{||c c c c c||} 
 \hline
 Market & Lang. & Intents & Train(ST) & Test(MT) \\ [0.5ex] 
 \hline\hline
 BR & pt & 316 & 66k & 372 \\ 
 ID & id & 481 & 161k & 1145 \\
 MY & en,ms & 473 & 74k & 1417 \\
 PH & en,fil & 237 & 33k & 189 \\
 SG & en & 360 & 76k & 737 \\
 TH & th & 359 & 60k & 502 \\
 TW & zh-tw & 373 & 31k & 353 \\
 VN & vi & 389 & 178k & 525 \\ [1ex]
 \hline
\end{tabular}}
\caption{The major languages, number of intents, and the number of samples in each market for Single Turn (ST) and Multi-Turn (MT). }
\label{table:data_train_test}
\end{table}

\section{Prompt Demonstration}
\label{appendix:c}

\subsection{Prompt for ICL ($\mathcal{P}$)}
To fit the width of this paper, we use SQ to represent Similar Question.
\begin{prompt}[title={Prompt for ICL ($\mathcal{P}$)},mathescape=true]
# Task Description
A chat between a curious user and an artificial intelligence assistant. The assistant gives helpful, detailed, and polite answers to the user's questions. USER: Determine the intent for the targetted message from the examples, you must use the context in the history messages to arrive at the best answer. 
# Examples
[Content] SQ_1 [Intent] Intent_name_1
[Content] SQ_2 [Intent] Intent_name_2
[Content] SQ_3 [Intent] Intent_name_3

# Note
DO NOT create new intent on your own, you must strictly use the intents in the examples.
DO NOT provide any explanation.
Output ONLY ONE intent for the targgetted message.
Consider the context from previous messages if the targetted message is unclear.

# Context
message 1: User's query
message 2: User's query with Entity
[Content] Last user's query

# Output
ASSISTANT: [Intent] <$Model$ $generated$ $Intent\_name$>
\end{prompt}

\subsection{Prompt for ICL ($\mathcal{P}_{symbolic}$)}
In $\mathcal{P}_{symbolic}$ the original label name $l$ of each intent in $\mathcal{E}_{symbolic}$ are replaced with single-token symbols, e.g. `A', `B', ..., which bear no meaning to the intents they represented. Explanation will be made in the instruction prompt $\mathcal{T}_{symbolic}$ to link the symbols back to their original intent label $y_j$, and the model is instructed to generated the symbols instead of full label names. 

\begin{prompt}[title={Prompt for ICL ($\mathcal{P}_{symbolic}$)},mathescape=true]
# Task Description
Content is Same as $\mathcal{P}$

# Examples
[Content] SQ_1 [Intent] A
[Content] SQ_1 [Intent] B
<$omitted$>
[Content] SQ_1 [Intent] B

# Intent options
A is Intent_name_1
B is Intent_name_2

# Note
Content is Same as $\mathcal{P}$

# Context
Format is same as $\mathcal{P}$

# Output
ASSISTANT: [Intent]
\end{prompt}

\subsection{$\mathcal{P}_{prepend}$}
In $\mathcal{P}_{prepend}$, representative symbols for each intent will be prepend to the original label name $l$, such that they are separated by an extra character as boundary, e.g. label ``logistics$>$how long will it take to receive order?" will be represented as ``A$>$logistics$>$how long will it take to receive order?". Note that the instruction prompt $\mathcal{T}$ remains the same, the trick is to limit the model generation token count to 1 on API level. 

\begin{prompt}[title={Prompt for ICL ($\mathcal{P}_{prepend}$)},mathescape=true]
# Task Description
Content is Same as $\mathcal{P}$

# Examples
[Content] SQ_1 [Intent] A>Intent_name_1
[Content] SQ_2 [Intent] B>Intent_name_2
<$omitted$>
[Content] SQ_3 [Intent] B>Intent_name_2

# Note
Content is Same as $\mathcal{P}$

# Context
Format is same as $\mathcal{P}$

# Output
ASSISTANT: [Intent] B
\end{prompt}

\subsection{$\mathcal{P}_{formatted}$}
\begin{prompt}[title={Prompt for ICL ($\mathcal{P}_{formatted}$)},mathescape=true]
# Task Description
Content is Same as $\mathcal{P}$

# Examples
Format is same as $\mathcal{P}_{prepend}$

# Note
Content is Same as $\mathcal{P}_{prepend}$

# Context
[History msg 1] Query
[History msg 2] Query with Entity
[Content] that is the order id 

# Output
ASSISTANT: [Intent]
\end{prompt}

\section{More Light-weight Deployment Method}
\label{apendix:d}

The multi-component architecture can be complicated to implement for real-time systems. Alternatively, this method can be used offline as a multi-turn data pseudo-labeling tool to train a classification model. The training method will be the same as the $\mathcal{M}_c$ classifier in the paper, just with pseudo-labeled multi-turn data added to the original data with only single-turn samples. We also did experiment to ensure the quality of the model trained pseudo-labelled data.

The prompt used for the experiment is $\mathcal{P}_{formatted}$. Since this is not a real-time task, and we don’t need to care about the pipeline response time, we also did self-consistency checking on the LLM outputs to ensure the quality of pseudo-labels. For this checking, the in-context learning part is run three times per sample, with the in-context examples sorted in three fashions according to their scores: ascending, descending, and random. 70k of online chat logs are sampled for pseudo-labelling, and only those having consistent labels after 3 runs will be kept for training. Around 12\% of the data will yield inconsistent results and be discarded. We validated that doing self-consistency this way can improve the average accuracy by \textbf{4.48\%} (from 64.64\% to 69.12\%), and thus the quality of the pseudo-label.

Moreover, the classifier trained with the high-quality multi-turn data generated by our pipeline can achieve better overall performance than the best original proposed method by \textbf{1.89\%} (64.64\% vs 66.53\%). This is all while cutting the deployment cost to just one classical classification model, but with the trade off of offline training time.


%% file: acl.bbl
\begin{thebibliography}{24}
\providecommand{\natexlab}[1]{#1}

\bibitem[{Brown et~al.(2020)Brown, Mann, Ryder, Subbiah, Kaplan, Dhariwal, Neelakantan, Shyam, Sastry, Askell, Agarwal, Herbert-Voss, Krueger, Henighan, Child, Ramesh, Ziegler, Wu, Winter, Hesse, Chen, Sigler, Litwin, Gray, Chess, Clark, Berner, McCandlish, Radford, Sutskever, and Amodei}]{gpt3}
Tom~B. Brown, Benjamin Mann, Nick Ryder, Melanie Subbiah, Jared Kaplan, Prafulla Dhariwal, Arvind Neelakantan, Pranav Shyam, Girish Sastry, Amanda Askell, Sandhini Agarwal, Ariel Herbert-Voss, Gretchen Krueger, Tom Henighan, Rewon Child, Aditya Ramesh, Daniel~M. Ziegler, Jeff Wu, Clemens Winter, Christopher Hesse, Mark Chen, Eric Sigler, Mateusz Litwin, Scott Gray, Benjamin Chess, Jack Clark, Christopher Berner, Sam McCandlish, Alec Radford, Ilya Sutskever, and Dario Amodei. 2020.
\newblock \href {https://api.semanticscholar.org/CorpusID:218971783} {Language models are few-shot learners}.
\newblock \emph{ArXiv}, abs/2005.14165.

\bibitem[{Budzianowski et~al.(2018)Budzianowski, Wen, Tseng, Casanueva, Ultes, Ramadan, and Ga{\v{s}}i{\'c}}]{budzianowski2018multiwoz}
Pawe{\l} Budzianowski, Tsung-Hsien Wen, Bo-Hsiang Tseng, Inigo Casanueva, Stefan Ultes, Osman Ramadan, and Milica Ga{\v{s}}i{\'c}. 2018.
\newblock Multiwoz--a large-scale multi-domain wizard-of-oz dataset for task-oriented dialogue modelling.
\newblock \emph{arXiv preprint arXiv:1810.00278}.

\bibitem[{Ghosal et~al.(2021)Ghosal, Majumder, Mihalcea, and Poria}]{ghosal2021exploring}
Deepanway Ghosal, Navonil Majumder, Rada Mihalcea, and Soujanya Poria. 2021.
\newblock Exploring the role of context in utterance-level emotion, act and intent classification in conversations: An empirical study.
\newblock In \emph{Findings of the Association for Computational Linguistics: ACL-IJCNLP 2021}, pages 1435--1449.

\bibitem[{Halder et~al.(2024)Halder, Lim, Chan, and Zhang}]{halder2024survey}
Sajal Halder, Kwan~Hui Lim, Jeffrey Chan, and Xiuzhen Zhang. 2024.
\newblock A survey on personalized itinerary recommendation: From optimisation to deep learning.
\newblock \emph{Applied Soft Computing}, 152:111200.

\bibitem[{Lee and Lee(2022)}]{lee2022compm}
Joosung Lee and Wooin Lee. 2022.
\newblock Compm: Context modeling with speaker’s pre-trained memory tracking for emotion recognition in conversation.
\newblock In \emph{Proceedings of the 2022 Conference of the North American Chapter of the Association for Computational Linguistics: Human Language Technologies}, pages 5669--5679.

\bibitem[{Liu et~al.(2021)Liu, Shen, Zhang, Dolan, Carin, and Chen}]{liu2021makes}
Jiachang Liu, Dinghan Shen, Yizhe Zhang, Bill Dolan, Lawrence Carin, and Weizhu Chen. 2021.
\newblock \href {https://arxiv.org/abs/2101.06804} {What makes good in-context examples for gpt-$3$?}
\newblock \emph{Preprint}, arXiv:2101.06804.

\bibitem[{Liu et~al.(2022)Liu, Ng, Gui, Singhal, Blessing, Wood, and Lim}]{liu2022title2vec}
Junhua Liu, Yung~Chuen Ng, Zitong Gui, Trisha Singhal, Lucienne~TM Blessing, Kristin~L Wood, and Kwan~Hui Lim. 2022.
\newblock Title2vec: A contextual job title embedding for occupational named entity recognition and other applications.
\newblock \emph{Journal of Big Data}, 9(1):99.

\bibitem[{Mo et~al.(2023)Mo, Nie, Huang, Mao, Zhu, Li, and Liu}]{Mo2023LearningTR}
Fengran Mo, Jianyun Nie, Kaiyu Huang, Kelong Mao, Yutao Zhu, Peng Li, and Yang Liu. 2023.
\newblock \href {https://api.semanticscholar.org/CorpusID:259076134} {Learning to relate to previous turns in conversational search}.
\newblock \emph{Proceedings of the 29th ACM SIGKDD Conference on Knowledge Discovery and Data Mining}.

\bibitem[{Qin et~al.(2023)Qin, Wu, Zhang, Li, Luan, Wang, Wang, and Cui}]{qin2023bert}
Xiangyu Qin, Zhiyu Wu, Tingting Zhang, Yanran Li, Jian Luan, Bin Wang, Li~Wang, and Jinshi Cui. 2023.
\newblock Bert-erc: Fine-tuning bert is enough for emotion recognition in conversation.
\newblock In \emph{Proceedings of the AAAI Conference on Artificial Intelligence}, volume~37, pages 13492--13500.

\bibitem[{Qu et~al.(2019)Qu, Yang, Croft, Zhang, Trippas, and Qiu}]{Qu_2019}
Chen Qu, Liu Yang, W.~Bruce Croft, Yongfeng Zhang, Johanne~R. Trippas, and Minghui Qiu. 2019.
\newblock \href {https://doi.org/10.1145/3295750.3298924} {User intent prediction in information-seeking conversations}.
\newblock In \emph{Proceedings of the 2019 Conference on Human Information Interaction and Retrieval}, CHIIR ’19. ACM.

\bibitem[{Reimers and Gurevych(2019)}]{reimers-2019-sentence-bert}
Nils Reimers and Iryna Gurevych. 2019.
\newblock \href {https://arxiv.org/abs/1908.10084} {Sentence-bert: Sentence embeddings using siamese bert-networks}.
\newblock In \emph{Proceedings of the 2019 Conference on Empirical Methods in Natural Language Processing}. Association for Computational Linguistics.

\bibitem[{Ren and Xue(2020)}]{9082602}
Fuji Ren and Siyuan Xue. 2020.
\newblock \href {https://doi.org/10.1109/ACCESS.2020.2991484} {Intention detection based on siamese neural network with triplet loss}.
\newblock \emph{IEEE Access}, 8:82242--82254.

\bibitem[{Rojas et~al.(2020)Rojas, Bustamante, Cabezudo, and Oncevay}]{Rojas2020EfficientSF}
Kervy~Rivas Rojas, Gina Bustamante, Marco Antonio~Sobrevilla Cabezudo, and Arturo Oncevay. 2020.
\newblock \href {https://api.semanticscholar.org/CorpusID:218517070} {Efficient strategies for hierarchical text classification: External knowledge and auxiliary tasks}.
\newblock \emph{ArXiv}, abs/2005.02473.

\bibitem[{Rubin et~al.(2021)Rubin, Herzig, and Berant}]{Rubin2021LearningTR}
Ohad Rubin, Jonathan Herzig, and Jonathan Berant. 2021.
\newblock \href {https://api.semanticscholar.org/CorpusID:245218561} {Learning to retrieve prompts for in-context learning}.
\newblock \emph{ArXiv}, abs/2112.08633.

\bibitem[{Shen et~al.(2021)Shen, Wu, Yang, and Quan}]{shen2021directed}
Weizhou Shen, Siyue Wu, Yunyi Yang, and Xiaojun Quan. 2021.
\newblock Directed acyclic graph network for conversational emotion recognition.
\newblock In \emph{Proceedings of the 59th Annual Meeting of the Association for Computational Linguistics and the 11th International Joint Conference on Natural Language Processing (Volume 1: Long Papers)}, pages 1551--1560.

\bibitem[{Shin et~al.(2021)Shin, Lin, Thomson, Chen, Roy, Platanios, Pauls, Klein, Eisner, and Durme}]{Shin2021ConstrainedLM}
Richard Shin, C.~H. Lin, Sam Thomson, Charles~C. Chen, Subhro Roy, Emmanouil~Antonios Platanios, Adam Pauls, Dan Klein, Jas' Eisner, and Benjamin~Van Durme. 2021.
\newblock \href {https://api.semanticscholar.org/CorpusID:233297024} {Constrained language models yield few-shot semantic parsers}.
\newblock \emph{ArXiv}, abs/2104.08768.

\bibitem[{Weld et~al.(2021)Weld, Huang, Long, Poon, and Han}]{weld2021survey}
H.~Weld, X.~Huang, S.~Long, J.~Poon, and S.~C. Han. 2021.
\newblock \href {https://arxiv.org/abs/2101.08091} {A survey of joint intent detection and slot-filling models in natural language understanding}.
\newblock \emph{Preprint}, arXiv:2101.08091.

\bibitem[{Wu et~al.(2021{\natexlab{a}})Wu, Su, and Juang}]{wu2021contextaware}
Ting-Wei Wu, Ruolin Su, and Biing-Hwang Juang. 2021{\natexlab{a}}.
\newblock \href {https://arxiv.org/abs/2109.01267} {A context-aware hierarchical bert fusion network for multi-turn dialog act detection}.
\newblock \emph{Preprint}, arXiv:2109.01267.

\bibitem[{Wu et~al.(2021{\natexlab{b}})Wu, Lu, Hajishirzi, and Ostendorf}]{Wu2021DIALKIKI}
Zeqiu Wu, Bo-Ru Lu, Hannaneh Hajishirzi, and Mari Ostendorf. 2021{\natexlab{b}}.
\newblock \href {https://api.semanticscholar.org/CorpusID:237485380} {Dialki: Knowledge identification in conversational systems through dialogue-document contextualization}.
\newblock In \emph{Conference on Empirical Methods in Natural Language Processing}.

\bibitem[{Xu and Sarikaya(2014)}]{6853573}
Puyang Xu and Ruhi Sarikaya. 2014.
\newblock \href {https://doi.org/10.1109/ICASSP.2014.6853573} {Contextual domain classification in spoken language understanding systems using recurrent neural network}.
\newblock In \emph{2014 IEEE International Conference on Acoustics, Speech and Signal Processing (ICASSP)}, pages 136--140.

\bibitem[{Yu et~al.(2021)Yu, He, Zhang, Du, Pasupat, and Li}]{Yu2021FewshotIC}
Dian Yu, Luheng He, Yuan Zhang, Xinya Du, Panupong Pasupat, and Qi~Li. 2021.
\newblock \href {https://api.semanticscholar.org/CorpusID:233219405} {Few-shot intent classification and slot filling with retrieved examples}.
\newblock In \emph{North American Chapter of the Association for Computational Linguistics}.

\bibitem[{Yuan et~al.(2019)Yuan, jie Zhou, Li, Lv, Zhu, Han, and Hu}]{Yuan2019MultihopSN}
Chunyuan Yuan, Wen jie Zhou, Mingming Li, Shangwen Lv, Fuqing Zhu, Jizhong Han, and Songlin Hu. 2019.
\newblock \href {https://api.semanticscholar.org/CorpusID:202776649} {Multi-hop selector network for multi-turn response selection in retrieval-based chatbots}.
\newblock In \emph{Conference on Empirical Methods in Natural Language Processing}.

\bibitem[{Zhang et~al.(2020)Zhang, Ye, Zhang, Qiu, Fu, Li, Yang, and Sun}]{zhang2020multi}
Jinghan Zhang, Yuxiao Ye, Yue Zhang, Likun Qiu, Bin Fu, Yang Li, Zhenglu Yang, and Jian Sun. 2020.
\newblock Multi-point semantic representation for intent classification.
\newblock In \emph{Proceedings of the AAAI Conference on Artificial Intelligence}, volume~34, pages 9531--9538.

\bibitem[{Zhu et~al.(2023)Zhu, Zhang, Huang, Wu, and Xu}]{Zhu2023HiTINHT}
He~Zhu, Chong Zhang, Junjie Huang, Junran Wu, and Ke~Xu. 2023.
\newblock \href {https://api.semanticscholar.org/CorpusID:258865236} {Hitin: Hierarchy-aware tree isomorphism network for hierarchical text classification}.
\newblock In \emph{Annual Meeting of the Association for Computational Linguistics}.

\end{thebibliography}
